\documentclass{article}
\usepackage{spconf,amsmath,graphicx,hyperref}

\usepackage{multirow}  
\usepackage{booktabs}  
\usepackage{algorithm} 
\usepackage{algpseudocode} 
\usepackage{amssymb}
\usepackage{multirow}
\usepackage{enumitem}
\setlist[itemize]{leftmargin=3.5mm}
\DeclareMathOperator{\Tr}{Tr}

\title{ScoreNF: Score-Based Normalizing Flows for Sampling Unnormalized Distributions}
\name{Vikas Kanaujia, Vipul Arora}
\address{
Department of Electrical Engineering, IIT Kanpur, India}
\begin{document}
\ninept
\maketitle
\begin{abstract}
Unnormalized probability distributions are central to modeling complex physical systems across various scientific domains. Traditional sampling methods, such as Markov Chain Monte Carlo (MCMC), often suffer from slow convergence, critical slowing down, poor mode mixing, and high autocorrelation. In contrast, likelihood-based and adversarial machine learning models, though effective, are heavily data-driven, requiring large datasets and often encountering mode covering and mode collapse. In this work, we propose ScoreNF, a score-based learning framework built on the Normalizing Flow (NF) architecture, integrated with an Independent Metropolis-Hastings (IMH) module, enabling efficient and unbiased sampling from unnormalized target distributions. We show that ScoreNF maintains high performance even with small training ensembles, thereby reducing reliance on computationally expensive MCMC-generated training data. We also present a method for assessing mode-covering and mode-collapse behaviours. We validate our method on synthetic 2D distributions (MOG-4 and MOG-8) and the high-dimensional $\phi^4$ lattice field theory distribution, demonstrating its effectiveness for sampling tasks.

\end{abstract}
\section{Introduction}
\label{sec:intro}
Unnormalized probability distributions—such as the Boltzmann distribution of form $e^{-H(x)}$ are extensively encountered in a wide range of scientific fields, including statistical physics, biological sciences and Bayesian inference \cite{Akhound-SadeghR24,albergo2019flow,li2018neural,bosese}. These distributions often arise in scenarios where the partition function or normalization constant is analytically intractable, particularly in high-dimensional settings. Efficient sampling from such distributions is crucial for estimating physical observables, performing uncertainty quantification, and enabling posterior inference. However, traditional sampling methods like Markov Chain Monte Carlo (MCMC) \cite{besag2004introduction} face significant limitations, including slow convergence, critical slowing down, poor mode mixing, and high autocorrelation between samples. 

In recent years, machine learning-based approaches have shown considerable promise in addressing these limitations. Several generative models like Normalising flows \cite{NF_survey,papamakarios2021normalizing,rezende2015variational}, GAN \cite{goodfellow2014generative,gui2021review}, VAE \cite{kingma2019introduction,Kingma2013AutoEncodingVB} and Diffusion models \cite{ho2020denoising} have been proposed  for approximating target distributions.  By leveraging neural networks, these methods can learn complex transformations that map simple base distributions to the target distribution, enabling efficient sampling. 

In the context of un-normalized distributions, Boltzmann Generators \cite{noe2019boltzmann} leveraging Normalizing Flows are commonly used to generate independent, unbiased samples from the Boltzmann distribution by learning invertible mappings from a simple latent space to the complex target distribution. Despite their strengths, Boltzmann Generators present several limitations. Their training typically depends on MCMC-generated samples to estimate KL divergence, reintroducing significant computational overhead. Moreover, the use of the forward KL divergence (FKL) as an objective induces mode-covering behavior—i.e., assigning probability mass to low-density regions and resulting in high variance in downstream observable estimates. In contrast, optimizing with reverse KL divergence (RKL) tends to induce mode collapse, as it emphasizes only a few of the high-density regions while neglecting low-probability modes, thereby failing to capture the full support of the target distribution and leading to biased estimates of observables.

Similarly other machine learning models based on likelihood maximization or adversarial training offer alternative solutions, they typically require large datasets and often suffer from issues such as mode collapse and mode covering. This motivates the development of more efficient and reliable sampling frameworks for unnormalized distributions, particularly in data-scarce or computationally demanding regimes.

Score-based generative models \cite{song2019generative}, that aim to learn the score function \( \nabla_x \log p_d(x) \)—the gradient of the data log-density—have been widely applied in various domains such as image generation \cite{mcsm}, audio generation \cite{pascual2023full,chenwavegrad}, and more. Once trained, these models enable sampling via Langevin dynamics, where the learned score function—estimated through score matching—guides the iterative generation of samples from the target distribution. 
In this paper, we incorporate the learning philosophy of score estimation into the Normalizing Flow (NF) framework. Unlike traditional score-based models that rely on iterative Langevin dynamics for sampling, our approach enables sample generation through a single-pass transformation of noise, leveraging the invertibility of the flow architecture.
Our main contributions are as follows:


\begin{itemize}
\item We propose ScoreNF, a framework that integrates score-based learning into normalizing flows (NFs) to model the target distribution more effectively. By leveraging score information, ScoreNF enhances the expressiveness of the model, resulting in improved approximation fidelity and mitigating common issues such as mode collapse and mode coverage.
\item We demonstrate that our proposed method, ScoreNF, maintains near-identical performance even with a significantly reduced ensemble size during training. This substantially lowers the reliance on computationally expensive MCMC simulations for data generation.
\item We assess the efficacy of the proposed method across several benchmark distributions, including MOG-4, MOG-8, and the 64-dimensional scalar $\phi^4$ theory distribution. 
\item Additionally, we present an evaluation method to assess the extent of mode coverage and mode collapse in the modelled distribution.
\end{itemize}
The codes are made available  \href{https://github.com/madhavlab/2026_scorenf}{here} at this repository.

\section{Method}
\label{sec:method}
\subsection{Problem Formulation}
Given a dataset of samples $\mathcal{D} = \{\mathbf{x}_i \in \mathcal{X}\}_{i=1}^N$ drawn from a target distribution $p(\mathbf{x})$, where $\mathbf{x} \in \mathbb{R}^d$, the objective is to learn a parameterized generative model $q_\theta(\mathbf{x})$ such that $q_\theta(\mathbf{x}) \approx p(\mathbf{x})$. The goal is to achieve a high-fidelity approximation of the target distribution while utilizing as few samples as possible in the learning.

\subsection{Background}
Here, we briefly review the key principles of score-based generative models and normalizing flows relevant to our proposed approach.

\textbf{Score-based generative models} \cite{song2019generative,song2021scorebased} aim to approximate an unknown data distribution \( p(\mathbf{x}) \), \( \mathbf{x} \in \mathbb{R}^d \), by learning its score function, defined as the gradient of the log-density, \( \nabla_{\mathbf{x}} \log p(\mathbf{x}) \).


Rather than modeling density $p(\mathbf{x})$ directly, these methods learn a parametric score function $s_\theta(\mathbf{x}) \approx \nabla_{\mathbf{x}} \log p(\mathbf{x})$ estimated by minimizing the Fisher divergence between the true and model scores:

\begin{align}
   \mathcal{L}(\theta) = \frac{1}{2}\mathbb{E}_{p(\mathbf{x})} \left[ \left\| s_\theta(\mathbf{x}) - \nabla_{\mathbf{x}} \log p(\mathbf{x}) \right\|^2 \right] 
\end{align}

In the absence of true score $s(\mathbf{x}) = \nabla_{\mathbf{x}} \log p(\mathbf{x})$, $\mathcal{L}(\theta)$ can be simplified under certain regularity conditions \cite{hyvarinen05a}  as
\begin{align}
    \mathcal{L}(\theta) = \mathbb{E}_{p(\mathbf{x})} [\text{Tr}(\nabla_x s_\theta(\mathbf{x})) +\frac{1}{2}\|s_\theta(\mathbf x)\|^2] 
\end{align}
where $\Tr$ is the trace. It could also be approximated using denoising score matching (DSM)\cite{song2019generative}. Once the score function is learned, new samples can be generated by simulating a reverse-time stochastic differential equation (SDE) or Langevin dynamics \cite{song2019generative,song2021scorebased}.

\textbf{Normalising flows} \cite{NF_survey,papamakarios2021normalizing} construct complex probability distributions by applying a sequence of invertible transformations to a simple base distribution. Given a base random variable $\mathbf{z} \sim p_{\mathbf{z}}(\mathbf{z})$, a normalizing flow defines a bijective mapping $\mathbf{x} = f_\theta(\mathbf{z})$, where $f_\theta$ is an invertible function with tractable Jacobian determinant.

The probability density of $\mathbf{x}$ under the model $q_\theta(\mathbf{x})$ is obtained using the change of variables formula:

\begin{align}
    q_\theta(\mathbf{x}) = p_{\mathbf{z}}(f_\theta^{-1}(\mathbf{x})) \left| \det \left( \frac{\partial f_\theta^{-1}(\mathbf{x})}{\partial \mathbf{x}} \right) \right|
\label{density_equation}
\end{align}


%
By designing $f_\theta$ such that both the inverse and the Jacobian determinant are tractable, normalizing flows allow for exact likelihood estimation and efficient sampling. 

\textbf{Independent Metropolis Hastings (IMH) Algorithm:}
The Metropolis-Hastings (MH) algorithm is a Markov Chain Monte Carlo (MCMC) method for sampling from a target distribution $p(x)$, known either exactly or up to a normalizing constant. It constructs a Markov chain that satisfies the detailed balance condition \cite{chib1995understanding}. A new sample $\mathbf{x'}$ is proposed from a distribution $q(\mathbf{x'}|\mathbf{x})$, and accepted with probability:
\begin{align}
    p_{\text{accept}}(\mathbf{x'}|\mathbf{x}) = \min\left(1, \frac{q(\mathbf{x}|\mathbf{x'}) p(\mathbf{x'})}{q(\mathbf{x'}|\mathbf{x}) p(\mathbf{x})} \right)
\end{align}

In the Independent Metropolis-Hastings (IMH) variant \cite{brofos2022adaptation}, proposals are drawn independently of the current state, i.e., $q(\mathbf{x'}|\mathbf{x}) = q(\mathbf{x'})$, simplifying the acceptance probability to:
\begin{align}
\label{acceptance_prob}
    p_{\text{accept}}(\mathbf{x'}|\mathbf{x}) = \min\left(1, \frac{q(\mathbf{x}) p(\mathbf{x'})}{q(\mathbf{x'}) p(\mathbf{x})} \right)
\end{align}
This is useful when sampling from $p(\mathbf{x})$ is intractable but its unnormalized density is accessible.

\begin{figure}[t] 
    \centering
    \includegraphics[width=\linewidth]{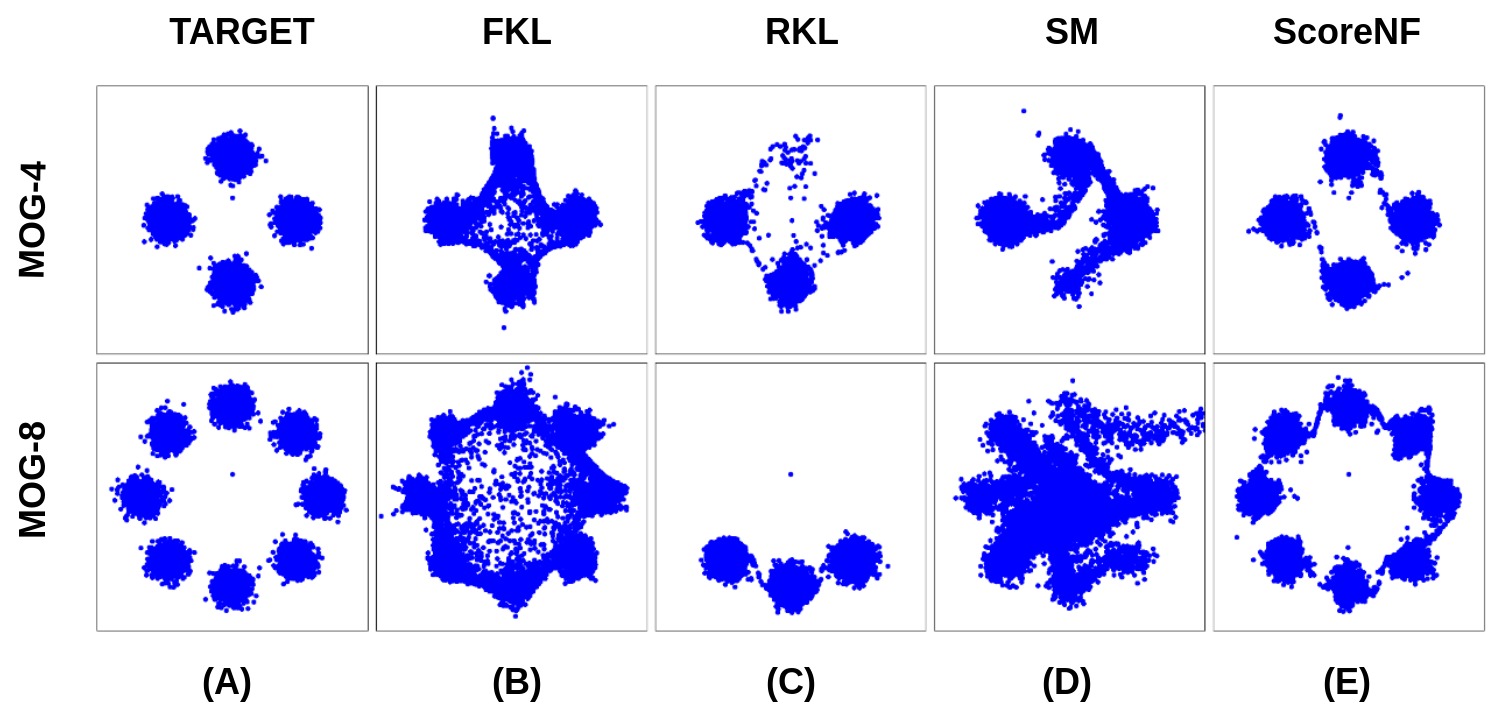}
    \vspace{-5mm}
    \caption{Sample plots for the MOG-4 and MOG-8 distributions generated using FKL, RKL, SM, and ScoreNF. Mode collapse is evident for the RKL approach, while mode covering behavior is observed in both FKL and SM approaches across all synthetic distributions.}
    \label{fig:sample_plot}
    \vspace{-2.5mm}
\end{figure}

\begin{table*}[ht]
    \centering
    \scalebox{0.8}{
    \begin{tabular}{l|cccc|cccc}
    \toprule
     & \multicolumn{4}{c}{MOG-4} & \multicolumn{4}{c}{MOG-8}  
    \\\cmidrule(lr){2-5}\cmidrule(lr){6-9}
     & NLL$\downarrow$ & RNLL($\downarrow$) & ESS($\uparrow$) & AR$\uparrow$ & NLL$\downarrow$ & RNLL($\downarrow$) & ESS($\uparrow$) & AR$\uparrow$\\\midrule
    FKL   & 2.53 $\pm$ 0.10 & 3.54 $\pm$ 0.89 & 50.76 $\pm$ 23.77 & 73.28 $\pm$ 0.73 & 3.30 $\pm$ 0.04 & 4.21 $\pm$ 0.45 & 45.37 $\pm$ 21.30 & 62.31 $\pm$ 1.00 \\
    RKL   & 3.00 $\pm$ 0.64 & \textbf{2.21 $\pm$ 0.02} & 18.73 $\pm$ 5.40 & 72.42 $\pm$ 5.89 & 372.84 $\pm$ 232.36 & \textbf{2.94 $\pm$ 0.13} & \textbf{78.83 $\pm$ 24.17} & \textbf{92.77 $\pm$ 0.44}\\
    SM    & 2.94 $\pm$ 0.05 & 3.33 $\pm$ 0.10 & 33.72 $\pm$ 1.96 & 46.24 $\pm$ 1.56 & 4.72 $\pm$ 0.32 & 14.04 $\pm$ 2.91 & 34.41 $\pm$ 13.83 & 18.35 $\pm$ 5.15\\
    ScoreNF & \textbf{2.21 $\pm$ 0.01} & 2.27 $\pm$ 0.02 & \textbf{83.06 $\pm$ 24.82} & \textbf{88.75 $\pm$ 0.42} & \textbf{3.01 $\pm$ 0.02} & 3.07 $\pm$ 0.06 & 48.08 $\pm$ 21.86 & 77.51 $\pm$ 0.84\\
    \bottomrule
    \end{tabular}
    }
    \vspace{-1mm}
    \caption{Results for the MOG-4 and MOG-8 distributions using 1,000 training samples. Reported values represent the average and standard error computed over three random seeds for each method.}
    \label{tab:mog4_8 results}
\end{table*}

\begin{table}[t]
    \vskip -0.4 cm
    \centering
    \scalebox{0.8}{
    \begin{tabular}{l|c|c|c|c}
    \toprule
          & NLL($\downarrow$) & RNLL($\downarrow$) & ESS($\uparrow$) & AR(\%)($\uparrow$)\\
    \midrule
    FKL   & 13.26 $\pm$ 0.10  & -10.35 $\pm$ 0.98 &27.76 $\pm$ 7.24  & 19.06 $\pm$ 0.97\\ 
    RKL   & 22.63 $\pm$ 3.85  & -18.93 $\pm$ 0.97&41.76 $\pm$ 0.99  & 68.29 $\pm$ 3.55\\
    SM    & 16.31 $\pm$ 0.10  &\textbf{-20.74 $\pm$ 0.12} & 33.77 $\pm$ 12.96 & \textbf{68.41 $\pm$ 1.76} \\
    ScoreNF & \textbf{12.05 $\pm$ 0.02} & -16.31 $\pm$ 0.21& \textbf{45.52 $\pm$ 0.41 } & 55.75 $\pm$ 0.55\\ 
    \bottomrule
    \end{tabular}
    }
    \vspace{-1.5mm}
    \caption{Results for $\phi^4$ model distribution (d = 64) using 10,000 training samples. Reported values represent the average and standard error computed over three random seeds for each method.}
    \vskip -0.15 cm
    \label{tab:phi4_results}
\end{table}
\subsection{Proposed Method}
In the  NF framework, training is typically performed by minimizing either the forward KL divergence (FKL), i.e.,  $KL(p||q_{\theta})$, or the reverse KL divergence (RKL), i.e., $KL(q_{\theta}||p)$, where $p(\mathbf{x})$ denotes the target distribution and 
$q_{\theta}(\mathbf{x})$, the modelled distribution parameterized by the flow parameters, $\theta$. Optimization using FKL requires samples from the target distribution, as it relies on maximizing likelihood. This approach tends to induce a mode-covering behavior in the learned distribution, encouraging coverage of low probability  regions with significant probability mass, as illustrated in Fig.~\ref{fig:sample_plot}(B).
In contrast, RKL-based training requires access only to an explicit form of the target density function; it does not require samples from the target distribution.  Training samples are drawn from the base distribution and transformed through the flow model. However, RKL minimization typically exhibits a mode-seeking tendency, often leading to mode collapse where the learned distribution fails to capture all modes of the target, as illustrated in Fig.~\ref{fig:sample_plot}(C). To address the limitations associated with both mode-collapse (arising from RKL minimization) and mode-covering (from FKL minimization), we leverage a score-matching objective to train NF. RKL optimization tends to concentrate on a few dominant modes. In contrast, score-based learning utilizes target samples as anchor points for matching local gradient estimates, which promotes more balanced coverage across all modes. Score learning, when combined with RKL, facilitates the learning of all modes in the target distribution, thereby reducing mode collapse. 

Incorporating score-based learning within the NF framework requires computation of both the target score function, $s(\mathbf{x})$, and the model score, $s_{\theta}(\mathbf{x})$.
Given that the target density $p(\mathbf{x})$ is either known exactly or even up to normalizing constant, the exact score function can be evaluated analytically, given by:
\begin{align}
    s(\mathbf{x}) &= \nabla_{\mathbf{x}}\log p(\mathbf{x}) \\
                  &= \nabla_{\mathbf{x}}\log \Tilde{p}(\mathbf{x}), \quad p(\mathbf{x}) \propto \Tilde{p}(\mathbf{x})\\
                  &= -\nabla_{\mathbf{x}} H(\mathbf{x}), \quad \Tilde{p}(\mathbf{x})= e^{-H(\mathbf{x})}
\end{align}
While prior approaches \cite{hyvarinen05a,song2019generative} directly parameterize the score function using neural networks, NFs explicitly model the density (Eq.~\ref{density_equation}). This enables direct computation of the model score via automatic differentiation, $s_{\theta}(\mathbf{x}) = \nabla_{\mathbf{x}} \log q_{\theta}(\mathbf{x})$. The score matching objective simplifies to:
\begin{align}
    \mathcal{L}_{SM}(\theta) &= \mathbb{E}_{\mathbf{x} \sim p(\mathbf{x})}\lVert \nabla_{\mathbf{x}}\log q_{\theta}(\mathbf{x}) - \nabla_{\mathbf{x}}\log p(\mathbf{x}) \rVert^2 \\
    &= \mathbb{E}_{\mathbf{x} \sim p(\mathbf{x})}\lVert \nabla_{\mathbf{x}}\log q_{\theta}(\mathbf{x}) + \nabla_{\mathbf{x}} H(\mathbf{x}) \rVert^2
\label{SM_equation}
\end{align}

Score-based training alone is insufficient to fully capture the target distribution, as it primarily aligns local gradient information rather than global density structure. To enhance distribution learning, we incorporate both the RKL and a score-matching objective into the training loss. The overall training objective is defined as:
\begin{align}
    \mathcal{L}_{net}(\theta) = \mathcal{L}_{RKL}(\theta) + \lambda_{1} \mathcal{L}_{SM}(\theta)
\end{align}
Where $\lambda_1$ is a hyperparameter, $\mathcal{L}_{SM}$ is defined in Eq.~\ref{SM_equation}, and $\mathcal{L}_{RKL}$ is defined as follows:
{\small
\begin{align}
    \mathcal{L}_{\text{RKL}} = \mathbb{E}_{\mathbf{z}} \left[ \log p_z(\mathbf{z}) - \log \left| \det \frac{\partial f_\theta(\mathbf{z})}{\partial \mathbf{z}}\right| - \log p(f_\theta(\mathbf{z})) \right]
\end{align}
}
Score matching aligns the score functions—i.e., the gradients of log-densities—of the model and target, its behaviour encourages broader mode coverage \cite{lyu2009interpretation}. In contrast, the RKL objective inherently exhibits a mode-seeking bias, which can lead to mode collapse. To mitigate this issue, we introduce the score-matching loss as a regularizer to the RKL loss. This combination helps preserve multiple modes during optimization. As training progresses and the model becomes more aligned with the target distribution, the weight of the score-matching loss is gradually annealed down, allowing the RKL term to refine the density estimation without inducing mode-collapse.

One may wonder if score matching here could be replaced by FKL, which is also sample driven.
However, these desired effects are not observed with FKL. This discrepancy arises from the fundamental differences between score learning and FKL optimization. FKL, primarily based on likelihood estimation, depends exclusively on samples and not on the target density $p(\mathbf x)$ itself, whereas score learning uses both the samples as well as $p(\mathbf x)$.
As a result, spurious samples have little impact on score learning but can significantly distort the distribution learned with FKL. Even with limited samples, score learning effectively guides the modeled distribution toward the target. We observe the same empirically as well.

After training, the NF model can generate samples and evaluate their exact densities. To correct for sampling bias introduced by model approximation, we apply the IMH algorithm, using the learnt distribution as the proposal. Samples are then accepted or rejected based on the acceptance probability defined in Eq.~\ref{acceptance_prob}, yielding asymptotically unbiased samples from the target distribution.

\section{Experiments}
\label{sec:exp_results}
This section outlines the experiments conducted on various distributions, including brief descriptions of each distribution and the evaluation metrics used.
\subsection{Distributions}
\textbf{MOG-4 and MOG-8:}
To assess model performance, we use synthetic 2D Mixture of Gaussians distributions—MOG-4 and MOG-8 \cite{kanaujia2024advnf}. These benchmarks provide clear insights into mode coverage and mode collapse, as the ground truth distribution and its individual modes are known in advance. MOG-4 consists of a mixture of 4 Gaussian components, whereas MOG-8 contains 8 components.
Using the PyTorch distribution library, we generate 10,000 samples for training and 5,000 samples each for validation and testing, corresponding to $N=4,8$.

\textbf{Scalar $\phi^4$ theory:}
It is a computational physics model to study scalar field theories \cite{singha2023conditional}. Given a scalar field $\mathbf{x} \in \mathbb{R}^d$ defined on a 2D square lattice with $d$ sites, the energy function, $H(\mathbf{x})$ is given by:
\begin{align}
  H(\mathbf{x})=\sum_{l=1}^d \left( \lambda_1 x_l^4 + \lambda_2 x_l^2 + 2\sum_{l'\in n(l)} (x_l^2 - x_l x_{l'}) \right) 
\label{eq:H_phi4}
\end{align}
where $n(l)$ denotes the set of lattice sites adjacent to the  $l^{th}$ site. The coupling constants are set to $\lambda_1 = 4$ \& $\lambda_2 = -4$. HMC is employed to generate 10,000 samples for training, and 5,000 samples each for validation and testing, all corresponding to the system size $d = 64$. 

\begin{table*}[!t]
    \centering
    \scalebox{0.8}{
    \begin{tabular}{l|l|cccc|cccc}
    \toprule
    
     & &\multicolumn{4}{|c|}{\textbf{MOG-4}} & \multicolumn{4}{|c}{\textbf{MOG-8}}  
    \\\cmidrule(lr){1-6}\cmidrule(lr){7-10}
    Sample Size & Model & NLL$\downarrow$ & RNLL($\downarrow$) & ESS($\uparrow$) & AR$\uparrow$ & NLL$\downarrow$ & RNLL($\downarrow$) & ESS($\uparrow$) & AR$\uparrow$\\\midrule
    \multirow{2}{*}{10000} & FKL & 2.20 $\pm$ 0.01 & 2.38 $\pm$ 0.09 & 89.74 $\pm$ 6.20 & 84.81 $\pm$ 0.36 & 2.92 $\pm$ 0.00 & 3.06 $\pm$ 0.04 & 74.89 $\pm$ 17.64 & 82.29 $\pm$ 0.36\\
    & ScoreNF & 2.18 $\pm$ 0.01 & 2.26 $\pm$ 0.02 & 96.22 $\pm$ 3.21  & 90.38 $\pm$ 0.52 & 2.94 $\pm$ 0.01 & 3.00 $\pm$ 0.05 & 68.26 $\pm$ 21.44 & 80.10 $\pm$ 0.71\\
    \hline
    \multirow{2}{*}{1000}  & FKL &2.53 $\pm$ 0.10 & 3.54 $\pm$ 0.89 & 50.76 $\pm$ 23.77 & 73.28 $\pm$ 0.73 & 3.30 $\pm$ 0.04 & 4.21 $\pm$ 0.45 & 45.37 $\pm$ 21.30 & 62.31 $\pm$ 1.00\\
    & ScoreNF & 2.21 $\pm$ 0.01 & 2.27 $\pm$ 0.02 & 83.06 $\pm$ 24.82 & 88.75 $\pm$ 0.42 & 3.01 $\pm$ 0.02 & 3.07 $\pm$ 0.06 & 48.08 $\pm$ 21.86 & 77.51 $\pm$ 0.84\\
    \hline
    \multirow{2}{*}{250}  & FKL & 3.20 $\pm$ 0.13 & 7.08 $\pm$ 0.38 & 54.43 $\pm$ 24.10 & 34.84 $\pm$ 0.96 & 3.91 $\pm$ 0.06 & 6.86 $\pm$ 0.91 & 41.43 $\pm$ 17.73 & 32.32 $\pm$ 0.63\\
     &ScoreNF & 2.24 $\pm$ 0.02 & 2.27 $\pm$ 0.04 & 70.06 $\pm$ 23.15 & 84.50 $\pm$ 0.69 & 3.11 $\pm$ 0.04 & 3.14 $\pm$ 0.13 & 43.62 $\pm$ 25.68 & 77.67 $\pm$ 1.66\\   
    \bottomrule
    \end{tabular}
    }
    \caption{Performance comparison of the ScoreNF model with FKL across varying training sample sizes on the MOG-4 and MOG-8 distributions.}
    \label{tab:ensemble_effect_mog}
\end{table*}

\subsection{Metrics}
We use the following four metrics for evaluation:

\textbf{Assessing Mode Collapse and Mode Coverage:}
Generally, \textbf{Negative Log-Likelihood (NLL)} is used to measures how well the model fits the data.
\begin{align}
    \text{NLL} = - \mathbb{E}_{p(\mathbf{x})}[\log(q_{\theta}(\mathbf{x}))]
\end{align}
A high NLL suggests mode collapse. But generally, NLL values are not significantly affected in case of mode coverage. We present \textbf{reverse NLL (RNLL)} as a strong indicator of mode coverage.
\begin{align}
    \text{RNLL} = - \mathbb{E}_{q_{\theta}(\mathbf{x})} [\log p(\mathbf{x})]
\end{align}
Together with NLL, this metric diagnoses the distribution behavior. High NLL and low RNLL indicate mode collapse, low NLL and high RNLL indicate mode coverage, while low values for both indicate that the model closely matches the target distribution. 



\textbf{Effective Sample Size (ESS)} : 
It measures the quantity of independent information contained within a sample and is defined as
\begin{align}
    \text{ESS} = \frac{(\frac{1}{N}\sum_{i} p(\mathbf{x}_i)/q_{\theta}(\mathbf{x}_i))^2}{\frac{1}{N}\sum_i (p(\mathbf{x}_i)/q_{\theta}(\mathbf{x}_i))^2} 
\end{align}
Higher values indicate more effective sampling of the target distribution.

\textbf{Acceptance Rate (AR)} : 
It is the percentage of accepted samples out of all evaluated samples in the IMH algorithm \cite{kanaujia2024advnf}.

We compare the proposed method against three baselines: NF trained with forward KL divergence (FKL), NF trained with reverse KL divergence, and NF trained via score matching (SM).

\begin{table}[t]
    \vskip -0.4cm
    \centering
    \scalebox{0.7}{
    \begin{tabular}{l|l|c|c|c|c}
    \toprule
    Sample Size & Model & NLL($\downarrow$) & RNLL($\downarrow$) & ESS($\uparrow$) & AR$\uparrow$ \\\midrule
    \multirow{2}{*}{10000} & FKL & 13.26 $\pm$ 0.10 & -10.35 $\pm$ 0.98 & 27.76 $\pm$ 7.24 & 19.06 $\pm$ 0.97\\
    & ScoreNF & 12.05 $\pm$ 0.02 & -16.31 $\pm$ 0.21 & 45.52 $\pm$ 0.41 & 55.75 $\pm$ 0.55 \\
    \hline
    \multirow{2}{*}{1000} & FKL & 14.49 $\pm$ 0.18 & -7.65 $\pm$ 1.14 & 30.24 $\pm$ 3.85 & 13.52 $\pm$ 2.12\\
    & ScoreNF & 12.06 $\pm$ 0.00 & -16.65 $\pm$ 0.06 & 33.99 $\pm$ 2.65 & 56.41 $\pm$ 0.80\\
    \hline
    \multirow{2}{*}{250}& FKL & 17.84 $\pm$ 0.08 & -7.11 $\pm$ 1.32 & 23.74 $\pm$ 1.99 & 9.11 $\pm$ 1.06\\
    & ScoreNF & 12.09 $\pm$ 0.02 & -15.88 $\pm$ 0.06 & 34.22 $\pm$ 2.66 & 53.51 $\pm$ 1.33\\
    \bottomrule
    \end{tabular}
    }
    \vspace{-2.5mm}  
    \caption{Performance comparison of the ScoreNF model with FKL across varying training sample sizes on the Scalar $\phi^4$ theory distribution.}
    \vskip -0.15cm
    \label{tab:ensemble_effect_phi4}
\end{table}

\section{Results and Discussion}
\label{sec:results}

\subsection{MOG-4 and MOG-8}
For the synthetic distributions, we compute all the evaluation metrics, the results of which are summarized in Table~\ref{tab:mog4_8 results}. A comparative visualization of the generated samples across all methods is provided in Fig.~\ref{fig:sample_plot}. We observe that FKL-based training exhibits mode-covering behaviour. In contrast, RKL-based training suffers from mode collapse, failing to capture one mode in the MOG-4 setting and up to five modes in MOG-8, which is further evidenced by increased NLL values. Despite these omissions, RKL may achieve better ESS, AR and RNLL scores due to its tendency to concentrate probability mass on fewer, high-density modes, thereby reducing penalization from uncovered regions. This is a limitation of the metrics, and not a merit of the method. Hence, NLL and RNLL should be seen together.
 
In contrast, our proposed method, ScoreNF, achieves superior performance, with sample plots closely aligning with the target distribution. This visual accuracy is corroborated by low NLL and RNLL values, reflecting better density estimation. 

\subsection{Scalar $\phi^4$ distribution}
The results for the scalar $\phi^4$ distribution are presented in Table \ref{tab:phi4_results}.
ScoreNF outperforms FKL-, RKL-, and SM-based methods, achieving low NLL values without mode collapse. Although RKL and SM exhibit low RNLL values indicative of limited mode coverage, their high NLL values reveal substantial mode collapse. The RNLL of ScoreNF is slightly higher than that of RKL and SM, indicating relatively high mode coverage, but it is still lower than that of FKL, reflecting relatively reduced mode coverage. ScoreNF achieves significantly better overall performance by avoiding mode collapse and mode covering present in other methods. 

Model performance with FKL, SM, and ScoreNF depends on the number of training samples from the target distribution. To assess this, we trained FKL and ScoreNF using ensembles of 10,000, 1,000, and 250 samples on both Synthetic and $\phi^4$ distributions. Results in Tables~\ref{tab:ensemble_effect_mog} and \ref{tab:ensemble_effect_phi4} show that FKL performance degrades notably as ensemble size decreases, consistent with its likelihood-based estimation \cite{braunstein1992large}. In contrast, ScoreNF remains robust and largely unaffected by reductions in ensemble size.

We also evaluate other baselines such as FAB~\cite{midgleyflow} and iDEM~\cite{Akhound-SadeghR24}, but exclude them from the main comparisons since they do not use target samples during training. Nonetheless, both methods are outperformed by ScoreNF on the scalar $\phi^4$ theory distribution (Table~\ref{phi4_additional_results}).

\begin{table}[t]
    \vskip -0.4cm
    \centering
    \scalebox{0.8}{
    \begin{tabular}{l|c|c|c}
    \toprule
          & NLL$\downarrow$ & RNLL$\downarrow$ & ESS(\%)$\uparrow$ \\
    \midrule
    FAB   & 13.48  & -4.31 & 7.0  \\ 
    iDEM   & 24.66 & -20.59 & 1.0 \\
    \hline
    ScoreNF & \textbf{12.05} & -16.31 & \textbf{45.52}\\ 
    \bottomrule
    \end{tabular}
    }
    \vspace{-2.5mm}
    \caption{Comparision of Results for $\phi^4$ model distribution  with other SOTA baselines.}
    \vskip -0.1 cm
    \label{phi4_additional_results}
\end{table}

Hybrid objectives for NFs have been widely studied, but often show limited robustness. The FKL–RKL hybrid~\cite{kanaujia2024advnf} performs poorly with small training ensembles, yielding high NLL and large variance in observables, while combining RKL with adversarial loss~\cite{kanaujia2024advnf} reduces mode collapse but still fails to capture all modes due to inherent limitations of adversarial training. 
In contrast, RKL with score matching (ScoreNF) performs more reliably, as score matching learns complementary information (gradient), promoting balanced mode coverage and effectively mitigates mode collapse. 

\section{Conclusions}
\label{sec:conc}
In this work, we present ScoreNF, a score-based learning method in the NF framework designed for sampling from unnormalized distributions. ScoreNF effectively addresses key challenges faced by traditional NF models, notably mitigating both mode collapse and mode covering. To assess mode collapse, negative log-likelihood (NLL) remains a reliable metric. To complement this, we use a metric, RNLL, which, when used alongside NLL, effectively quantifies the model’s mode coverage behavior. Combined, these metrics offer a robust framework for evaluating both mode collapse and mode coverage. ScoreNF, though reliant on target samples for score estimation, achieves strong performance even with limited data. Empirical results on the $\phi^4$ distribution and synthetic distributions (MOG-4 and MOG-8) demonstrate its ability to capture complex multimodal distributions. Future work may extend ScoreNF to larger systems with complex dynamics and multiple symmetries.



---------------------------------



\vfill\pagebreak

\bibliographystyle{IEEEbib}
\bibliography{strings,refs}

@inproceedings{Akhound-SadeghR24,
  author="Tara Akhound-Sadegh and Jarrid Rector-Brooks and Avishek Joey Bose and Sarthak Mittal and Pablo Lemos and Cheng-Hao Liu and Marcin Sendera and Siamak Ravanbakhsh and Gauthier Gidel and Yoshua Bengio and Nikolay Malkin and Alexander Tong",
  title="Iterated Denoising Energy Matching for Sampling from Boltzmann Densities",
  year="2024",
  booktitle="Proceedings of the 41st International Conference on Machine Learning",
  pages = "760-786",
  publisher = "PMLR"
}

@article{albergo2019flow,
  title={Flow-based generative models for Markov chain Monte Carlo in lattice field theory},
  author={Albergo, Michael S and Kanwar, Gurtej and Shanahan, Phiala E},
  journal={Physical Review D},
  volume={100},
  number={3},
  pages={034515},
  year={2019},
  publisher={APS}
}

@inproceedings{bosese,
  title={SE (3)-Stochastic Flow Matching for Protein Backbone Generation},
  author={Bose, Joey and Akhound-Sadegh, Tara and Huguet, Guillaume and FATRAS, Kilian and Rector-Brooks, Jarrid and Liu, Cheng-Hao and Nica, Andrei Cristian and Korablyov, Maksym and Bronstein, Michael M and Tong, Alexander},
  booktitle={The Twelfth International Conference on Learning Representations},
  year = {2023}
}

@article{li2018neural,
  title={Neural network renormalization group},
  author={Li, Shuo-Hui and Wang, Lei},
  journal={Physical review letters},
  volume={121},
  number={26},
  pages={260601},
  year={2018},
  publisher={APS}
}

@inproceedings{goodfellow2014generative,
  title={Generative Adversarial Nets},
  author={Goodfellow, Ian and Pouget-Abadie, Jean and Mirza, Mehdi and Xu, Bing and Warde-Farley, David and Ozair, Sherjil and Courville, Aaron and Bengio, Yoshua},
  booktitle={Advances in neural information processing systems},
  pages={2672--2680},
  year={2014}
}

@article{gui2021review,
  title={A review on generative adversarial networks: Algorithms, theory, and applications},
  author={Gui, Jie and Sun, Zhenan and Wen, Yonggang and Tao, Dacheng and Ye, Jieping},
  journal={IEEE transactions on knowledge and data engineering},
  volume={35},
  number={4},
  pages={3313--3332},
  year={2021},
  publisher={IEEE}
}

@article{Kingma2013AutoEncodingVB,
  title={Auto-Encoding Variational Bayes},
  author={Diederik P. Kingma and Max Welling},
  journal={CoRR},
  year={2013},
  volume={abs/1312.6114},
}

@article{kingma2019introduction,
  title={An introduction to variational autoencoders},
  author={Kingma, Diederik P and Welling, Max and others},
  journal={Foundations and Trends{\textregistered} in Machine Learning},
  volume={12},
  number={4},
  pages={307--392},
  year={2019},
  publisher={Now Publishers, Inc.}
}

@article{papamakarios2021normalizing,
  title={Normalizing flows for probabilistic modeling and inference},
  author={Papamakarios, George and Nalisnick, Eric and Rezende, Danilo Jimenez and Mohamed, Shakir and Lakshminarayanan, Balaji},
  journal={Journal of Machine Learning Research},
  volume={22},
  number={57},
  pages={1--64},
  year={2021}
}

@inproceedings{rezende2015variational,
  title={Variational inference with normalizing flows},
  author={Rezende, Danilo and Mohamed, Shakir},
  booktitle={International conference on machine learning},
  pages={1530--1538},
  year={2015},
  organization={PMLR}
}

@article{ho2020denoising,
  title={Denoising diffusion probabilistic models},
  author={Ho, Jonathan and Jain, Ajay and Abbeel, Pieter},
  journal={Advances in neural information processing systems},
  volume={33},
  pages={6840--6851},
  year={2020}
}

@article{singha2023conditional,
  title={Conditional normalizing flow for Markov chain Monte Carlo sampling in the critical region of lattice field theory},
  author={Singha, Ankur and Chakrabarti, Dipankar and Arora, Vipul},
  journal={Physical Review D},
  volume={107},
  number={1},
  pages={014512},
  year={2023},
  publisher={APS}
}

@article{kanaujia2024advnf,
  title={AdvNF: Reducing mode collapse in conditional normalising flows using adversarial learning},
  author={Kanaujia, Vikas and Scheurer, Mathias S and Arora, Vipul},
  journal={SciPost Physics},
  volume={16},
  number={5},
  pages={132},
  year={2024}
}

@article{hyvarinen05a,
  author  = {Aapo Hyv{{\"a}}rinen},
  title   = {Estimation of Non-Normalized Statistical Models by Score Matching},
  journal = {Journal of Machine Learning Research},
  year    = {2005},
  volume  = {6},
  number  = {24},
  pages   = {695--709}
}

@article{song2019generative,
  title={Generative modeling by estimating gradients of the data distribution},
  author={Song, Yang and Ermon, Stefano},
  journal={Advances in neural information processing systems},
  volume={32},
  year={2019}
}

@inproceedings{
song2021scorebased,
title={Score-Based Generative Modeling through Stochastic Differential Equations},
author={Yang Song and Jascha Sohl-Dickstein and Diederik P Kingma and Abhishek Kumar and Stefano Ermon and Ben Poole},
booktitle={International Conference on Learning Representations},
year={2021},
}

@inproceedings{brofos2022adaptation,
  title={Adaptation of the independent metropolis-hastings sampler with normalizing flow proposals},
  author={Brofos, James and Gabri{\'e}, Marylou and Brubaker, Marcus A and Lederman, Roy R},
  booktitle={International Conference on Artificial Intelligence and Statistics},
  pages={5949--5986},
  year={2022},
  organization={PMLR}
}

@incollection{besag2004introduction,
  title={An introduction to Markov chain Monte Carlo methods},
  author={Besag, Julian},
  booktitle={Mathematical foundations of speech and language processing},
  pages={247--270},
  year={2004},
  publisher={Springer}
}

@article{noe2019boltzmann,
  title={Boltzmann generators: Sampling equilibrium states of many-body systems with deep learning},
  author={No{\'e}, Frank and Olsson, Simon and K{\"o}hler, Jonas and Wu, Hao},
  journal={Science},
  volume={365},
  number={6457},
  pages={eaaw1147},
  year={2019},
  publisher={American Association for the Advancement of Science}
}

@ARTICLE{NF_survey,
  author={Kobyzev, Ivan and Prince, Simon J.D. and Brubaker, Marcus A.},
  journal={IEEE Transactions on Pattern Analysis and Machine Intelligence}, 
  title={Normalizing Flows: An Introduction and Review of Current Methods}, 
  year={2021},
  volume={43},
  number={11},
  pages={3964-3979},
}

@article{braunstein1992large,
  title={How large a sample is needed for the maximum likelihood estimator to be approximately Gaussian?},
  author={Braunstein, Samuel L},
  journal={Journal of Physics A: Mathematical and General},
  volume={25},
  number={13},
  pages={3813},
  year={1992},
  publisher={IOP Publishing}
}

@INPROCEEDINGS{mcsm,
  author={Shetty, Nishanth and Seelamantula, Chandra Sekhar},
  booktitle={ICASSP 2025 - 2025 IEEE International Conference on Acoustics, Speech and Signal Processing (ICASSP)}, 
  title={Monte Carlo Score Matching for Image Generation}, 
  year={2025},
  volume={},
  number={},
  pages={1-5},
  doi={10.1109/ICASSP49660.2025.10889041}}

@inproceedings{lyu2009interpretation,
  title={Interpretation and generalization of score matching},
  author={Lyu, Siwei},
  booktitle={Proceedings of the Twenty-Fifth Conference on Uncertainty in Artificial Intelligence},
  pages={359--366},
  year={2009}
}

@inproceedings{pascual2023full,
  title={Full-band general audio synthesis with score-based diffusion},
  author={Pascual, Santiago and Bhattacharya, Gautam and Yeh, Chunghsin and Pons, Jordi and Serr{\`a}, Joan},
  booktitle={ICASSP 2023-2023 IEEE International Conference on Acoustics, Speech and Signal Processing (ICASSP)},
  pages={1--5},
  year={2023},
  organization={IEEE}
}

@article{chib1995understanding,
  title={Understanding the metropolis-hastings algorithm},
  author={Chib, Siddhartha and Greenberg, Edward},
  journal={The american statistician},
  volume={49},
  number={4},
  pages={327--335},
  year={1995},
  publisher={Taylor \& Francis}
}

@inproceedings{chenwavegrad,
  title={WaveGrad: Estimating Gradients for Waveform Generation},
  author={Chen, Nanxin and Zhang, Yu and Zen, Heiga and Weiss, Ron J and Norouzi, Mohammad and Chan, William},
  booktitle={International Conference on Learning Representations},
  year = {2021}
}

@inproceedings{midgleyflow,
  title={Flow Annealed Importance Sampling Bootstrap},
  author={Midgley, Laurence Illing and Stimper, Vincent and Simm, Gregor NC and Sch{\"o}lkopf, Bernhard and Hern{\'a}ndez-Lobato, Jos{\'e} Miguel},
  booktitle={The Eleventh International Conference on Learning Representations},
  year ={2023}
}

\end{document}